\documentclass[a4paper,11pt]{article}
\usepackage{CLARIN2024}
\usepackage{times}
\usepackage{url}
\usepackage{latexsym}
\PassOptionsToPackage{hyphens}{url}\usepackage[hidelinks]{hyperref}
\usepackage{tabularx}
\usepackage{graphicx}
\usepackage{comment}
\usepackage[english]{babel}
\usepackage{csquotes}
\usepackage[
    backend=biber,
    style=apa,
    natbib=true,
    ]{biblatex}
\usepackage{covington} % if needed, for linguistic examples

\title{CLASSLA-Express: a Train of CLARIN.SI Workshops\\on Language Resources and Tools with Easily Expanding Route}
\author{Nikola Ljubešić and Taja Kuzman\\
  Department of Knowledge Technologies\\
  Jožef Stefan Institute\\
  Ljubljana, Slovenia \\
  {\tt \{nikola.ljubesic,taja.kuzman\}@ijs.si} \\\And % if needed: this makes a second column
  Ivana Filipović Petrović \\
  Linguistic Research Institute of the\\
  Croatian  Academy of Sciences and Arts\\
  Zagreb, Croatia\\
  {\tt ifilipovic@hazu.hr}
 \AND % if needed: this makes a second row
  Jelena Parizoska \\
  Faculty of Teacher Education\\
  University of Zagreb, Croatia\\
  {\tt jelena.parizoska@ufzg.unizg.hr} \\\And
  Petya Osenova \\
  AI and Language Technologies Division\\
  IICT-BAS \\
  Sofia, Bulgaria \\
 {\tt petya@bultreebank.org} \\
}

\date{}

\begin{document}
\maketitle
%The abstract should be a concise summary of the general thesis and conclusions of the paper. It should be no longer than 200 words.
\begin{abstract}
This paper introduces the CLASSLA-Express workshop series as an innovative approach to disseminating linguistic resources and infrastructure provided by the CLASSLA Knowledge Centre for South Slavic languages and the Slovenian CLARIN.SI infrastructure. The workshop series employs two key strategies: (1) conducting workshops directly in countries with interested audiences, and (2) designing the series for easy expansion to new venues. The first iteration of the CLASSLA-Express workshop series encompasses 6 workshops in 5 countries. Its goal is to share knowledge on the use of corpus querying tools, as well as the recently-released CLASSLA-web corpora -- the largest general corpora for South Slavic languages. In the paper, we present the design of the workshop series, its current scope and the effortless extensions of the workshop to new venues that are already in sight.
\end{abstract}

\section{Introduction} \label{intro}

% - - - - - - - IMPORTANT - - - - - - -
% The following footnote without marker is needed for the camera-ready
% version of the paper.
% Comment out the instructions (first text) and uncomment the 8 lines
% under "final paper" for your variant of English.
%
\blfootnote{
    %
    % for review submission
    %
    %\hspace{-0.65cm}  % space normally used by the marker
This work is licenced under a Creative Commons Attribution 4.0 International Licence. Licence details: \url{http://creativecommons.org/licenses/by/4.0/}
    % % final paper: en-uk version (to license, a licence)
    %
    % \hspace{-0.65cm}  % space normally used by the marker
    % This work is licensed under a Creative Commons
    % Attribution 4.0 International Licence.
    % Licence details:
    % {http://creativecommons.org/licenses/by/4.0/}
    %
    % % final paper: en-us version (to licence, a license)
    %
    % \hspace{-0.65cm}  % space normally used by the marker
    % This work is licenced under a Creative Commons
    % Attribution 4.0 International License.
    % License details:
    % {http://creativecommons.org/licenses/by/4.0/}
}

%\textbf{\emph{NOTE: Since reviewing will be double-blind, author names should not appear in any submission to be reviewed (but should be given in the final accepted paper)}}.Likewise, submissions to be reviewed must not mention contributors, project names, grant numbers, and names or URLs of resources or tools that have only been made publicly available in the last three weeks or are about to be made public.
% \textbf{\emph{NOTE: In the anonymous version submitted for initial review, your real names and addresses should \emph{not} be filled out here (see also the General Instructions section above).}}\\

% Avoid subsubsections.
% Citations within the text appear in parentheses as~\citep{Gusfield:97} or, if the author's name appears in the text itself, as \citet{Gusfield:97}.  

We present the CLASSLA-Express workshop series as an approach to promote CLARIN.SI\footnote{\url{https://www.clarin.si/info/about/}} linguistic resources and its infrastructure. This is achieved by 1) conducting workshops in countries with interested audiences, eliminating the need for individuals to travel to specific venues; 2) designing the workshops to be easily scalable to additional locations. By directly bringing workshops to countries with interested audiences and in some venues conducting the workshops in the national languages, a broader and more diverse group of individuals beyond the existing CLARIN.SI research community can be reached, including those who may face constraints in traveling or have a lower proficiency in English. This approach is not only more inclusive but also environmentally sustainable by only requiring the workshop lecturers to travel. Furthermore, the workshop series is structured in a way that facilitates seamless expansion to new locations. We share teaching materials, available in multiple languages, with interested colleagues from other institutions and nations, enabling the effortless extension of the workshop series to communities that were not initially targeted in the first iteration of CLASSLA-Express. Following the announcement of the first six workshops, researchers from Montenegro, Bosnia and Herzegovina, Croatia and Serbia have already expressed interest to take on the workshop, and the extension of the workshop series to additional countries and cities is already in sight.

 The first iteration of the CLASSLA-Express workshop series comprises six workshops that took place from April to September 2024 in the following cities: Zagreb (Croatia), Rijeka (Croatia), Belgrade (Serbia), Skopje (North Macedonia), Sofia (Bulgaria)\footnote{The workshop in Sofia is also supported by CLaDA-BG: {\url{https://clada-bg.eu/en/}}}, and Ljubljana (Slovenia). The main goal of the workshop is to familiarize the participants with the CLASSLA-web corpora \citep{ljubešić2024classlaweb}, the largest general corpora for South Slavic languages, as well as with other corpora and tools provided by the CLARIN.SI infrastructure, including the pipeline for linguistic annotation CLASSLA-Stanza \citep{ljubesic-dobrovoljc-2019-neural, terčon2023classlastanza} and concordancers (corpus querying tools). Of the six workshops, three are aimed at students of South Slavic languages, linguists and university lecturers and are held at universities which have Slavic departments. The other three workshops are held at international conferences on lexicography, language technologies and digital humanities. Those are aimed at linguists, lexicographers, computational linguists, corpus linguists and digital humanities scholars. 

In this extended abstract, we present the workshop scope and the topics it covers (Section \ref{sec:scope}), the first iteration of the CLASSLA-Express workshop series (Section \ref{sec:first-iteration}), the promotional strategies employed to engage workshop participants, such as enhancing brand visibility and disseminating information through various channels (Section \ref{sec:promotion}) and the workshop framework designed to facilitate expansion to additional locations (Section \ref{sec:scaling}).

\section{Goal and Content of the Workshops}
\label{sec:scope}

The CLASSLA-Express series of workshops aims to show participants how to use the CLASSLA-web corpora \citep{ljubešić2024classlaweb} in language research, but also language teaching, designing corpus-informed dictionaries and grammars, and in digital humanities. The CLASSLA-web corpora cover Slovenian, Croatian, Bosnian, Montenegrin, Serbian, Macedonian, and Bulgarian, which makes them the first collection of comparable corpora that cover the entire language group. The corpus collection comprises a total of 13 billion tokens of texts from 26 million documents and represents ones of the biggest openly-available corpora for each language. What is more, the Macedonian CLASSLA-web corpus is the first linguistically annotated corpus available for this language. The CLASSLA-web corpora are available for download at the CLARIN.SI repository\footnote{\url{https://www.clarin.si/repository/xmlui/}}, and for querying at the CLARIN.SI NoSketch Engine concordancer\footnote{\url{https://www.clarin.si/ske/}}. They are characterized by their recency, substantial size, and diverse composition encompassing various genres, registers, and topics. Consequently, they represent a highly valuable resource for conducting language analyses, developing lexicons, and facilitating other forms of linguistic research.

The main part of the workshop comprises hands-on exercises showing how to create queries in Croatian \citep{classla-web-hr}, Macedonian \citep{classla-web-mk}, Serbian \citep{classla-web-sr}, Bulgarian \citep{classla-web-bg}, and Slovenian \citep{classla-web-sl} web corpora to obtain data on meanings and uses of words, word forms, collocations and grammatical patterns.
The presented queries were carefully designed to cater for participants with various research interests: morphology (word forms), grammar (e.g., case, aspect, sentence patterns), lexicology (collocations, idiomatic expressions) and discourse analysis (uses of words and constructions in different types of texts). %A substantial number of queries are based on convenors’ own research in previous web corpora for South Slavic languages, e.g. hrWaC. 

In addition, as many corpus linguistic analyses are based on linguistic annotations of the corpora, e.g., part-of-speech tags, we familiarize the participants with the automatic language processing tool CLASSLA-Stanza \citep{ljubesic-dobrovoljc-2019-neural,terčon2023classlastanza} which was used to annotate the CLASSLA-web corpora. The tool is also accessible through a web platform\footnote{\url{https://clarin.si/oznacevalnik/eng/}} which allows researchers to explore it and evaluate its limitations that should be taken into account when conducting a linguistic study of automatically annotated data. Lastly, the CLASSLA-Express workshops introduce the participants to a wide range of language resources and technologies that are provided for their languages by the CLARIN.SI infrastructure, as well as the knowledge-sharing initiatives and support services offered by the CLARIN Knowledge Centre for South Slavic languages (CLASSLA)\footnote{\url{https://www.clarin.si/info/k-centre/}}.

\section{First Iteration of the CLASSLA-Express Workshop Series}
\label{sec:first-iteration}

The first iteration of the CLASSLA-Express workshops took place from April to September 2024 across five countries and six cities. These workshops were designed as half-day sessions, lasting approximately four hours each. Table \ref{workshop-details} provides information on the dates and locations of the workshops. The workshops were conducted in Croatian in Zagreb, Rijeka and Skopje, Serbian in Belgrade, Bulgarian in Sofia, and English in Ljubljana.

Each workshop was intended for approximately 20 participants, resulting in a total of over 100 participants. The number of participants was limited and the workshops were not provided in a hybrid manner, as their nature is highly interactive. % and hands-on, requiring active participation and real-time engagement from attendees.
Participants had varying levels of familiarity with CLARIN.SI and CLASSLA resources. %Some had no prior experience while others have some experience with other South Slavic corpora provided by CLARIN.SI. Before the workshop, participants predominantly used older web corpora for South Slavic languages which underscores the significance of the availability of such corpora for linguists.
After the workshop, the participants were surveyed regarding their intentions to use the CLASSLA-web corpora in their research endeavors. They intend to use the corpora for various research purposes, including collocation analysis, frequency analysis, lexicography, comparisons between languages, specialized language studies (pertaining to child language, loanwords, phraseology, gendered language and vulgarisms), as well as for natural language processing tasks. Additionally, some plan to incorporate corpora into their academic work, including teaching.
 
\begin{table}[htb]
%\begin{center}
%\begin{tabular}{|l|rl|}
\centering
\begin{tabularx}{\textwidth}{|m{0.10\linewidth}|m{0.20\linewidth}m{0.62\linewidth}|}
\hline
\bf Date & \bf Location & \bf Venue \\
\hline
19/4/2024 & Zagreb, Croatia & Faculty of Humanities and Social Sciences, University of Zagreb \\
26/4/2024 & Rijeka, Croatia & Center for Language Research, Faculty of Humanities and Social Sciences, University of Rijeka \\
29/5/2024 & Belgrade, Serbia & International conference Leksikografski susreti \\ %, Faculty of Philology, University of Belgrade\\
4/6/2024 & Skopje, North Macedonia & Blaže Koneski Faculty of Philology, Ss. Cyril and Methodius University \\
26/6/2024 & Sofia, Bulgaria & International CLaDA-BG Conference 2024 \\
18/9/2024 & Ljubljana, Slovenia & Language Technologies \& Digital Humanities Conference 2024\\
\hline
\end{tabularx}
%\end{tabular}
%\end{center}
\caption{\label{workshop-details} Details of the first iteration of the CLASSLA-Express workshops.}
\end{table}

\section{Promotional Activities}
\label{sec:promotion}

An important part of the organization is dissemination of information about the workshops. Firstly,  a dedicated web page\footnote{\url{https://www.clarin.si/info/k-centre/workshops/classla-express/}} was created and published within the CLARIN.SI website to centralize information about the workshops. The dissemination of the calls for participation involved both traditional academic channels, such as posting news items on various local and international mailing lists, as well as social media platforms, specifically the X\footnote{\url{https://twitter.com/ClarinSlovenia}} and LinkedIn\footnote{\url{https://www.linkedin.com/company/clarin-si}} profiles of CLARIN.SI. Leveraging social media enabled a wider outreach to individuals across diverse fields, expanding the dissemination beyond the linguistic community. A survey distributed among the participants after each workshop showed that most of them learned about the CLASSLA-Express workshop via mailing lists, however, many participants found out about the workshop by word of mouth or on social media.

After the workshops, we shared the key highlights from each event on the CLARIN.SI website\footnote{\url{https://www.clarin.si/info/k-centre/workshops/}} and on social media platforms.

\begin{figure}[htb]
	\centering
		\includegraphics[width=0.2\textwidth]{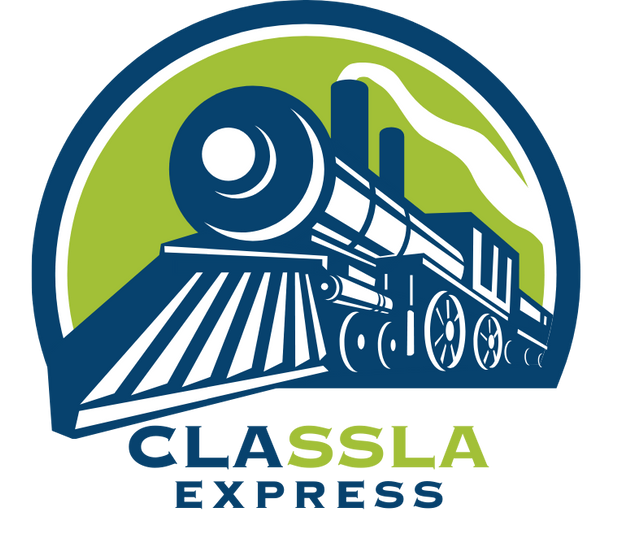}
    \caption{The CLASSLA-Express logo}
    \label{fig:logo}
\end{figure}

Moreover, efforts were made to enhance the workshops' visibility. We designed a logo for the workshop, shown in Figure \ref{fig:logo}. As an additional touch, certificates of attendance and logo-branded bags were prepared and distributed to participants.

\section{Extending the Workshop Series to New Venues}
\label{sec:scaling}

The first iteration of the CLASSLA-Express workshop series was mainly convened by the two lecturers who were willing to travel and conduct the workshop in five countries. However, the workshops can be easily extended to new locations in collaboration with colleagues at those venues, as was done with the organizer in Sofia, Bulgaria, a stop added after the first five stops were already agreed upon. 
We are prepared to offer assistance by providing templates for calls for participation, registration forms and certificates of attendance, as well as by promoting the workshops on the CLASSLA-Express website, and CLARIN.SI and CLASSLA communication platforms. Teaching materials have been developed in Croatian, Bulgarian and English and are available for sharing with interested partners. Given that CLASSLA-web corpora cover all South Slavic languages and have comparable format and scope, the queries designed for Slovenian, Croatian, Serbian, Macedonian and Bulgarian can be easily adapted to other South Slavic languages and applied to a specific CLASSLA-web corpus.

Following the announcement of the calls for participation for the initial CLASSLA-Express workshops, colleagues from Montenegro, and Bosnia and Herzegovina, as well as from additional locations in Croatia and Serbia, expressed interest in hosting workshops in their respective countries and cities, and these additional workshops are already in sight.

\section{Conclusion}

In this paper, we presented the strategy employed by the CLASSLA-Express workshop series to promote the CLARIN.SI linguistic resources and infrastructure, as well as the CLASSLA Knowledge Centre, beyond national borders. The workshops' approach involves directly engaging countries with interested audiences, eliminating the need for individuals to travel to a specific location. Additionally, in many locations, the workshops were conducted in the national languages. This method not only facilitates accessible and environmentally-friendly knowledge sharing but also broadens the reach of diverse individuals from different backgrounds and nations who become familiar with the resources and infrastructure. Furthermore, the dissemination of workshop information through various channels, including social media platforms, aims to engage individuals beyond the existing CLARIN.SI academic community, thereby expanding awareness of the resources and activities to a wider audience. Secondly, the workshop series is designed in such way that it can be easily extended to other locations. The CLASSLA-web corpora collection \citep{ljubešić2024classlaweb} which is the main focus of the workshops, comprises comparable corpora for all South Slavic languages. This allows the practical tasks to be adapted to any South Slavic CLASSLA-web corpus. In addition, we offer teaching materials in multiple languages to interested collaborators who wish to host the workshop in their own city. We also assist them with the logistics by providing templates for calls for participation, certificates of attendance, registration forms, and promoting the event through various channels.
 
The train of workshops has already gained momentum, with additional workshops in Montenegro, Bosnia and Herzegovina, Croatia and Serbia in sight. In the future, we plan to further extend the scope of the workshops by covering approaches on how to support corpus-based research with large language models, a discussion topic reappearing in each of the conducted workshops.  % to encompass even more cities and countries within the South Slavic region and beyond. 

\section*{Acknowledgments}
The CLASSLA-Express workshop series was funded by the CLARIN.SI infrastructure. The work in preparing and supporting the workshop series was partially funded by the programme P6-0411 ``Language Resources and Technologies for Slovene'' financed by the Slovenian Research and Innovation Agency (ARIS). The workshop in Sofia was also partially supported by CLaDA-BG, funded by the Ministry of Education and Science of Bulgaria.

\printbibliography

\begin{comment}
\section{Appendix}
Appendices, if any, directly follow the text and the
references (see above).  Letter them in sequence and provide an
informative title: {\bf Appendix A. Title of Appendix}.
\end{comment}

\end{document}